\documentclass{uber}
\usepackage{colortbl}
\usepackage{hyperref}
\usepackage{cleveref}
\usepackage{graphicx}
\usepackage{url}
\usepackage{booktabs}
\usepackage{multirow}
\usepackage{pifont}

\definecolor{gtgreen}{HTML}{39FF14}
\definecolor{predorange}{HTML}{FFB300}

\newcommand{\cmark}{\textcolor{green!70!black}{\ding{51}}}
\newcommand{\xmark}{\textcolor{red!70!black}{\ding{55}}}

\renewcommand\contributionformat[2][]{{\small $#1$\,#2}}

\title{GeoWAM: Visual Geometry World Action Models for Autonomous Driving}

\author{Yiren Lu$^{1,2}$}
\author{Xin Ye$^{1,\dagger}$}
\author{Jiaming Liu$^{1}$}
\author{Philip Jacobson$^{1}$}
\author{Jin Yao$^{1}$}
\author{Yi-chung Chen$^{1}$}
\author{Liam  Merino$^{1}$}
\author{Dhruva Dixith Kurra$^{1}$}
\author{Min Cai$^{1}$}
\author{Tom Lampo$^{1}$}
\author{Yu Yin$^{2,\dagger}$}
\author{Danhua Guo$^{1}$}
\author{Burhan Yaman$^{1\ddagger}$}
\affiliation[1]{Uber AV Labs}
\affiliation[2]{Case Western Reserve University}
\contribution[\dagger]{Corresponding authors}\contribution[\ddagger]{Project Lead}

\project{\url{https://yiren-lu.com/project_pages/geowam/}}
\date{\today}
\renewcommand{\uberlogo}{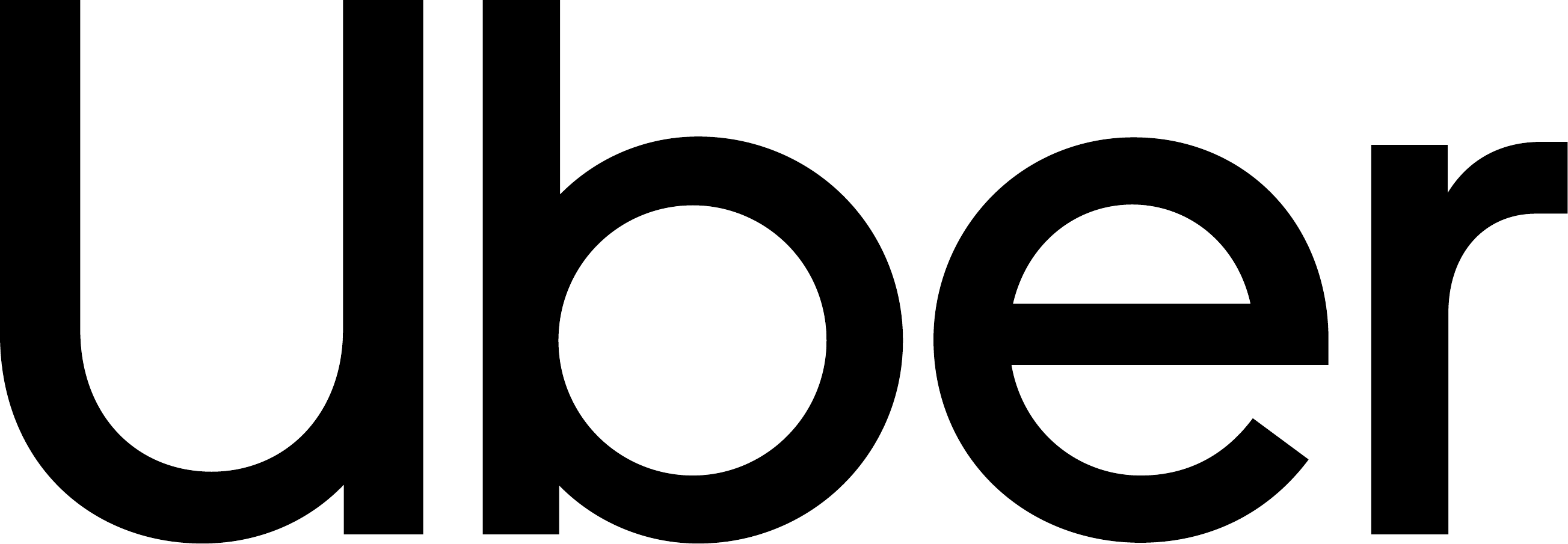}
\renewcommand{\uberlogosecond}{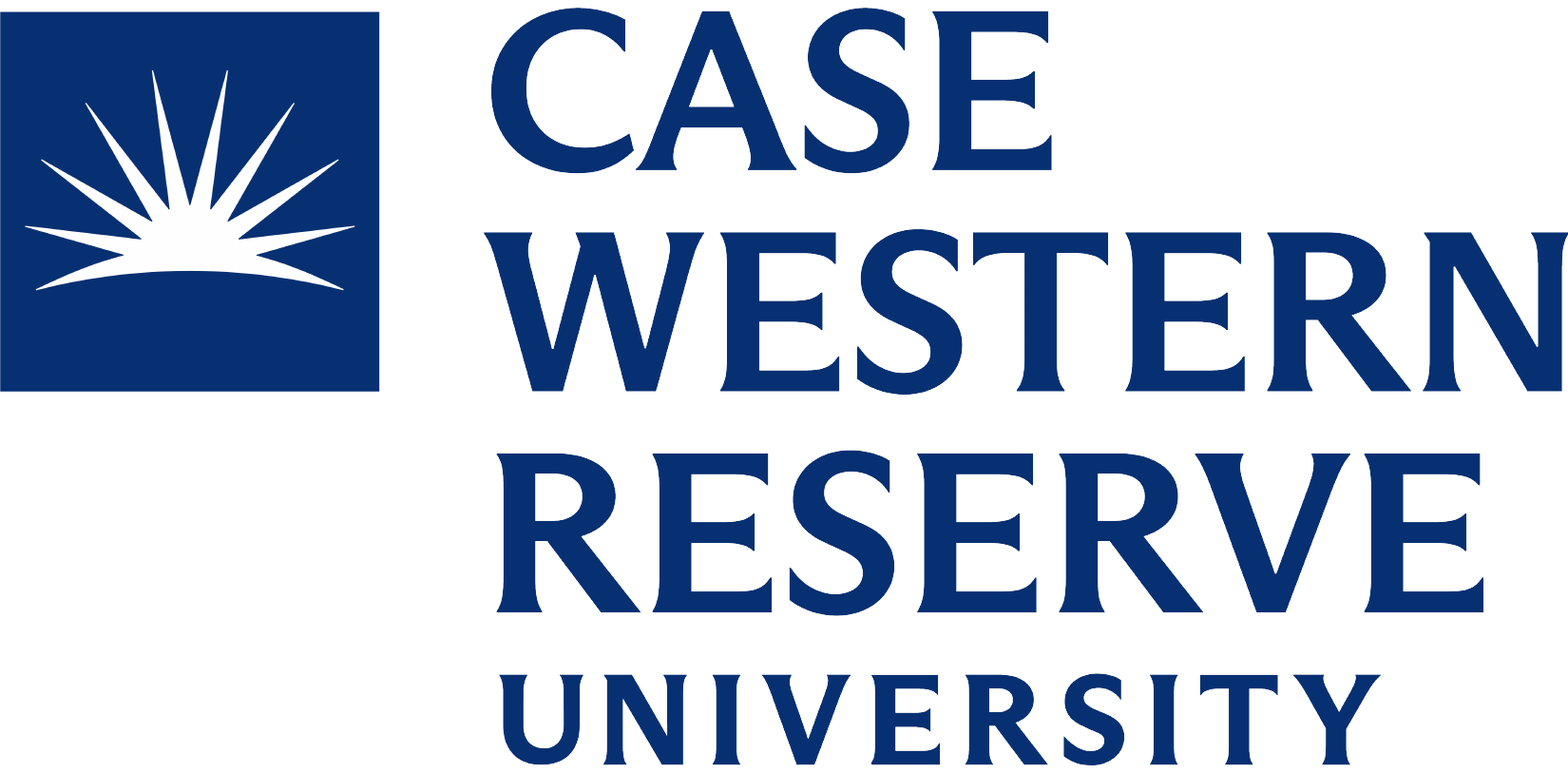}

\abstract{
World action models (WAMs) have recently gained increasing attention as a framework for jointly modeling scene evolution and ego actions in autonomous driving.
Most existing WAMs learn scene dynamics in pixel space by combining a video-generation backbone for future-observation prediction with an action head for ego-trajectory prediction.
Pixels, however, provide only an indirect representation of these dynamics: they entangle geometry and motion with appearance, texture, and illumination, forcing the model to infer three-dimensional transformations from two-dimensional observations.
We argue that point-based geometry provides a more natural state space for driving.
It explicitly captures spatial structure and both rigid and non-rigid scene dynamics while remaining aligned with the 3D space of driving actions.
Building on this insight, we introduce \textbf{GeoWAM}, a visual geometry world action model for autonomous driving.
Rather than predicting future images, GeoWAM is pretrained to forecast future scene geometry, yielding representations that jointly encode spatial structure and temporal evolution.
A geometry-conditioned action head then leverages these learned geometric dynamics to predict future ego-trajectories.
Extensive experiments show that GeoWAM outperforms image-based alternatives, achieving a combined EPDMS of 36.6 on \texttt{navhard} without PDMS supervision and strong zero-shot generalization to nuScenes, with a collision rate of 0.24\%. Scaling geometry pretraining with unlabeled data further improves performance, increasing the \texttt{navhard} score by 8.2\% to 39.6 and strengthening zero-shot transfer to nuScenes, where the collision rate is reduced by 50\% to 0.12\%. Together, these results establish geometry as an effective state representation for autonomous driving and geometry pretraining as a general, \textit{scalable} strategy for downstream planning.
}

\begin{document}
\maketitle

\section{Introduction}
\label{sec:intro}
Autonomous driving is fundamentally a problem of anticipation.
A capable driving agent must understand not only the current scene, but also how the scene is likely to evolve and how that evolution constrains its actions.
Recent end-to-end driving systems increasingly build on pretrained foundation models to acquire the broad representations needed for this task.

Vision-Language-Action (VLA) models transfer the semantic knowledge and reasoning capabilities of pretrained vision-language models to ego-trajectory prediction~\citep{hwang2024emma,zhou2026opendrivevla,tian2024drivevlm,jiang2026senna,wang2025alpamayo,li2026unidrivevla,yao2026vlga}.
While effective for high-level understanding and decision making, their action-centric objectives leave future scene dynamics implicit in the learned policy representation.

\begin{figure*}[htbp]
    \centering
    \includegraphics[width=1\linewidth]{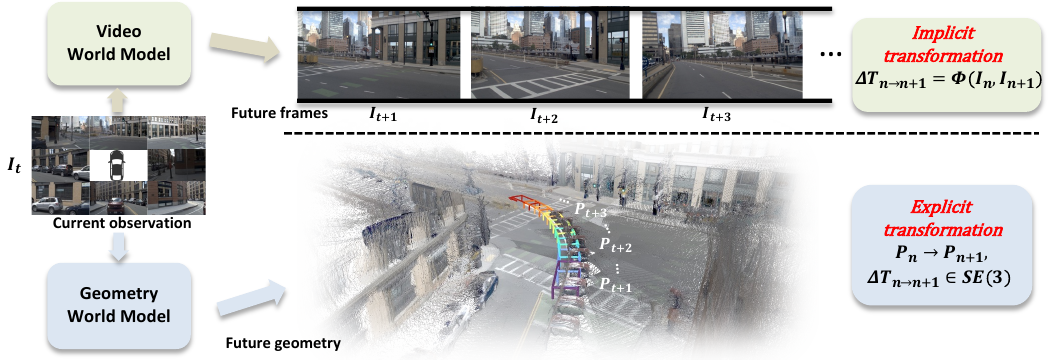}
    \caption{\textbf{Video-based versus geometry-based world modeling.}
    Video world models predict future pixels, from which the 3D transformations required for planning are  implicitly included and difficult to recover.
    Geometry world models instead predict explicit future 3D structure, whose temporal evolution directly exposes object and ego motion in a planning-aligned state space.}
    \label{fig:video_vs_geometry_world_model}
\end{figure*}

World models instead learn scene dynamics by forecasting future observations, often with high-capacity video-generation backbones~\citep{hu2023gaia,gao2024vista,wang2024driving,wang2024drivedreamer,hassan2025gem}.
World action models (WAMs) connect such prediction to planning by coupling future-state forecasting with trajectory generation~\citep{li2026drivevla,zheng2025world4drive,bartoccioni2025vavim,xiong2026unidrive}.
However, most WAMs inherit image-based state representations from video world models and therefore learn driving dynamics through future-frame prediction.

In this paper, we argue that pixels are not an ideal state representation for modeling driving dynamics.
Images encode geometry and motion \emph{indirectly}, entangling them with appearance, texture, and illumination.
As illustrated in \cref{fig:video_vs_geometry_world_model}, pixel-space predictions are not naturally aligned with trajectory planning in 3D, making the underlying 3D transformations difficult to recover.
Geometry, by contrast, provides a \emph{native} state space that directly represents 3D scene structure, while its temporal evolution reveals object and ego motion in a form naturally aligned with planning.

Building on this insight, we introduce \textbf{GeoWAM}, a visual geometry world action model for autonomous driving.
Instead of forecasting future images, GeoWAM is pretrained to predict future scene geometry, learning representations that jointly capture spatial structure and temporal evolution.
A geometry-conditioned action head then leverages these learned geometric dynamics to predict future ego trajectories.
By placing both world modeling and action prediction in a shared geometric space, GeoWAM provides the driving policy with representations explicitly organized around 3D structure and motion.
Moreover, we find that geometry pretraining is a data-scalable strategy that can improve downstream planning with increasing pretraining data while transferring effectively across datasets.
Extensive experiments on geometry prediction, open-loop and close-loop planning validate geometry as an effective state representation for modeling driving dynamics and supporting downstream planning.

In summary, our contributions are:
\begin{itemize}
  \item We introduce \textbf{GeoWAM}, a world action model built on geometry as a native state representation for driving. GeoWAM couples future-geometry modeling with trajectory prediction through an inverse-dynamics-like formulation, representing scene dynamics and ego trajectories within the same 3D coordinate space.
  \item We develop a data-scalable geometry pretraining strategy that learns spatial structure and temporal dynamics by forecasting future scene geometry from historical multiview observations. Scaling this pretraining with additional driving data consistently strengthens downstream planning and cross-dataset generalization.
  \item Extensive experiments on future-geometry prediction, planning under different settings showcase that GeoWAM consistently outperforms image-based alternatives, demonstrating the effectiveness of geometry as a state representation for autonomous driving.
\end{itemize}

\section{Related Work}
\label{sec:related_work}

\subsection{World Models and World Action Models for Autonomous Driving}
World models learn environment dynamics from past observations, in autonomous driving, they are predominantly formulated as future-video generation~\citep{hu2023gaia, wang2024drivedreamer, gao2024vista, hassan2025gem}.
Occupancy-based alternatives~\citep{zheng2024occworld, yang2025driving} forecast 3D scene evolution but require ground-truth occupancy supervision.

World action models (WAMs) couple future-state prediction with action generation, so that learned environment dynamics directly inform the policy.
Driving into the Future~\citep{wang2024driving} imagines multiview futures conditioned on candidate maneuvers and selects the trajectory whose predicted future receives the highest image-based reward.
VaViM and VaVAM~\citep{bartoccioni2025vavim} augment autoregressive video modeling with an action expert, while Epona~\citep{zhang2025epona} uses separate diffusion-based heads for image and trajectory prediction.
DriveVLA-W0~\citep{li2026drivevla} further uses future-image prediction as a dense self-supervised signal for driving vision-language-action models.
PWM~\citep{zhao2026forecasting} jointly forecasts state and action evolution, whereas DriveLaW~\citep{xia2026drivelaw} unifies planning and video generation within a shared latent driving world.
UniDrive-WM~\citep{xiong2026unidrive} unifies scene understanding, trajectory planning, and trajectory-conditioned future image generation within a shared vision-language architecture, whereas DriveWAM~\citep{shi2026drivewam} adapts a pretrained video diffusion transformer into a unified video-action policy under a joint flow-matching objective.
EponaV2~\citep{xu2026eponav2} additionally supervises future depth and semantic features alongside image prediction.
Despite their architectural differences, most driving WAMs use image or video generation as their world-modeling interface, leaving 3D scene evolution implicit.
GeoWAM makes metric geometry the primary prediction space and conditions trajectory generation on the resulting future geometric dynamics.

\subsection{Geometry Foundation Models}

General visual geometry models have increasingly replaced task-specific reconstruction pipelines with feed-forward prediction of dense geometric quantities.
DUSt3R~\citep{wang2024dust3r} introduced point-map regression for uncalibrated image pairs, removing the need to explicitly estimate camera parameters before reconstruction.
CUT3R~\citep{wang2025continuous} extends this formulation with a persistent state for continuous 3D perception, while VGGT~\citep{wang2025vggt} jointly infers camera parameters, depth, point maps, and tracks from a variable number of views.
MapAnything~\citep{11533250} further supports flexible geometric inputs and directly recovers metric-scale scene geometry in a unified feed-forward model.
Together, these methods establish dense point maps as a scalable representation for 3D reconstruction.

Driving scenes introduce additional requirements, including long-range metric accuracy, temporal motion, surround-view observations, and substantial variation across camera configurations.
DVGT~\citep{zuo2025dvgt} addresses these requirements with an ego-centric formulation that predicts metric point maps and ego poses from multi-frame, multi-view images using factorized spatial-temporal attention, while DVGT-2~\citep{zuo2026dvgt} scales this geometry representation toward autonomous-driving action modeling.
Most visual geometry models, however, primarily reconstruct the scene contained in observed images.
VGGT-World~\citep{sun2026vggtworld} is a recent exception that transforms VGGT into an autoregressive geometry world model for forecasting future geometry in its feature space.
GeoWAM instead forecasts dense metric point maps and connects the learned future geometric dynamics to downstream trajectory planning.

\section{Methodology}
\label{sec:methodology}

\begin{figure*}
    \centering
    \includegraphics[width=1\linewidth]{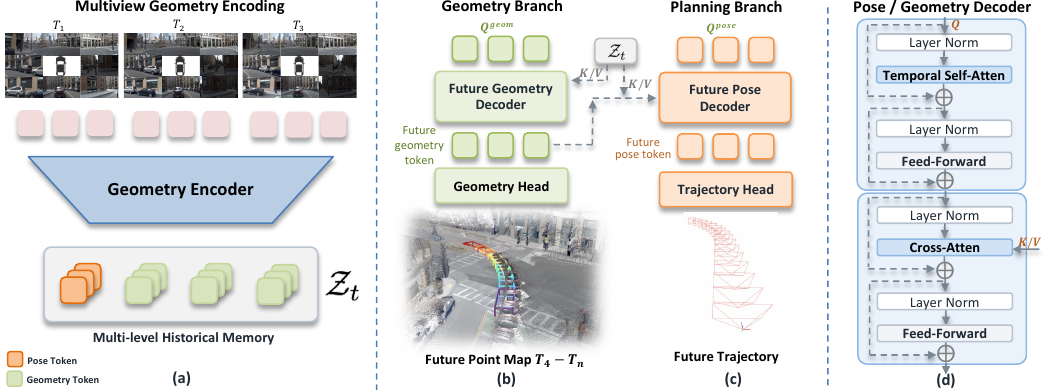}
    \caption{\textbf{GeoWAM pipeline.} (a) The geometry encoder builds a multi-level historical memory $\mathcal{Z}_t$ from multiview images. (b) Future-geometry queries attend to $\mathcal{Z}_t$ to produce future geometry tokens, which are then decoded into dense point maps. (c) Future-pose queries attend to the predicted geometry and $\mathcal{Z}_t$ to produce future pose tokens, which are then decoded into trajectories. (d) Detailed structure of pose/geometry decoder.}
    \label{fig:framework}
\end{figure*}

We introduce \textbf{GeoWAM}, a visual geometry world action model for autonomous driving, as illustrated in \cref{fig:framework}.
Unlike conventional video-based world action models,
GeoWAM first learns to forecast the three-dimensional evolution of a driving scene and then uses the predicted geometric dynamics to plan the future motion of the ego vehicle.
The training of GeoWAM consists of two stages.
In the first stage (\cref{sec:geometry_world_model}), we train a visual geometry world model using driving sequences to predict dense future scene geometry from historical multiview images.
In the second stage (\cref{sec:geowam_planning}), we extend the pretrained visual geometry world model with a geometry-conditioned action head and jointly finetune the complete model for both future-geometry prediction and trajectory planning.

\subsection{Visual Geometry World Model}
\label{sec:geometry_world_model}

\paragraph{Multiview geometry encoding.}
Let $\mathbf{I}_{t-K+1:t}=\{\mathbf{I}_{\tau}^{v}\mid \tau=t-K+1,\ldots,t;\ v=1,\ldots,V\}$ denote the $K$-frame multiview image history, where $\mathbf{I}_{\tau}^{v}\in\mathbb{R}^{3\times H\times W}$ is the image from camera $v$ at time $\tau$.
We adopt DVGT-2~\citep{zuo2026dvgt} as our geometry encoder $\mathcal{E}_{\theta}$ to encode the multiview image sequence into multi-level historical memory, as shown in \cref{fig:framework} (a).
We retain the outputs from $L$ selected feature levels, indexed by $\ell=1,\ldots,L$.
At each time step $\tau$ and feature level $\ell$, the encoder produces two types of tokens.
The geometry tokens $\mathbf{X}_{\tau}^{\ell}\in\mathbb{R}^{V\times P\times D}$ represent the spatial scene structure, while the ego tokens $\mathbf{E}_{\tau}^{\ell}\in\mathbb{R}^{V\times N_e\times D}$ encode ego-motion context.
Here, $P$ and $N_e$ are the number of geometry tokens and ego tokens per view, respectively, and $D$ is their feature dimension.
We concatenate them along the token dimension as
$\mathbf{Z}_{\tau}^{\ell}=[\mathbf{X}_{\tau}^{\ell};\mathbf{E}_{\tau}^{\ell}]\in\mathbb{R}^{V\times(P+N_e)\times D}$.
The complete historical geometry memory is defined as follows:
\begin{equation}
    \mathcal{Z}_{t}
    =
    \left\{
    \mathbf{Z}_{\tau}^{\ell}
    \mid
    \tau=t-K+1,\ldots,t;\ \ell=1,\ldots,L
    \right\}
    =
    \mathcal{E}_{\theta}(\mathbf{I}_{t-K+1:t}).
    \label{eq:geometry_encoding}
\end{equation}

\paragraph{Future geometry decoding.}
As shown in \cref{fig:framework} (b), the future-geometry branch predicts the scene over the next $F$ steps from a set of learned geometry queries.
Let $\mathbf{q}^{\mathrm{geom}}\in\mathbb{R}^{d}$ be a learned query seed, where $d$ is the hidden dimension of the future geometry decoder.
We replicate this seed for every future step, camera view, and spatial location.
The geometry query at future step $k$, view $v$, and spatial location $p$ is
\begin{equation}
    \mathbf{Q}_{t+k}^{\mathrm{geom},v,p}
    =
    \mathbf{q}^{\mathrm{geom}}
    +\mathbf{e}_{K+k}^{\mathrm{time}}
    +\mathbf{e}_{v}^{\mathrm{view}}
    +\mathbf{e}_{p}^{\mathrm{2D}},
    \label{eq:future_queries}
\end{equation}
where $\mathbf{e}_{K+k}^{\mathrm{time}}$ and $\mathbf{e}_{v}^{\mathrm{view}}$ are learned temporal and view embeddings, and $\mathbf{e}_{p}^{\mathrm{2D}}$ is a two-dimensional positional embedding.
We use $\mathcal{Q}_{t+1:t+F}^{\mathrm{geom}}=\{\mathbf{Q}_{t+k}^{\mathrm{geom},v,p}\}_{k,v,p}$ to denote all future geometry queries.

The decoder first projects the historical memory $\mathcal{Z}_{t}$ to its hidden dimension.
It then updates the geometry queries in two steps at each decoder layer, as shown in \cref{fig:framework} (d).
First, causal temporal self-attention models the evolution of each spatial location across the $F$ future steps.
Second, the updated queries cross-attend to $\mathcal{Z}_{t}$ to retrieve the relevant context from the memory.
After that, we obtain a future geometry latent $\hat{\mathbf{U}}_{t+k}\in\mathbb{R}^{V\times P\times d}$ for each future step.
For each feature level $\ell$, an output projection $\mathbf{W}_{\ell}\in\mathbb{R}^{D\times d}$ maps this latent to the feature space required by the geometry head:
\begin{equation}
    \begin{gathered}
        \hat{\mathcal{U}}_{t+1:t+F}
        =\mathcal{D}_{\phi}(\mathcal{Q}_{t+1:t+F}^{\mathrm{geom}},\mathcal{Z}_{t}),\\
        \hat{\mathbf{X}}_{t+k}^{\ell}
        =\hat{\mathbf{U}}_{t+k}\mathbf{W}_{\ell}^{\mathsf{T}},
        \quad k=1,\ldots,F,\ \ell=1,\ldots,L.
    \end{gathered}
    \label{eq:future_geometry_features}
\end{equation}
where $\hat{\mathcal{U}}_{t+1:t+F}=\{\hat{\mathbf{U}}_{t+k}\}_{k=1}^{F}$.
The level-specific outputs $\hat{\mathbf{X}}_{t+k}^{\ell}\in\mathbb{R}^{V\times P\times D}$ form a multi-level feature representation of the future scene.
The shared geometry head $\mathcal{G}_{\psi}$ decodes this representation into a dense point map and a per-pixel confidence map:
\begin{equation}
    \left(
    \hat{\mathbf{P}}_{t+k},
    \hat{\mathbf{C}}_{t+k}
    \right)
    =
    \mathcal{G}_{\psi}
    \left(
    \left\{\hat{\mathbf{X}}_{t+k}^{\ell}\right\}_{\ell=1}^{L}
    \right).
    \label{eq:future_pointmap}
\end{equation}
Here, $\hat{\mathbf{P}}_{t+k}\in\mathbb{R}^{V\times H\times W\times 3}$ stores one 3D point per image pixel in the ego coordinate system at time $t+k$, and $\hat{\mathbf{C}}_{t+k}\in\mathbb{R}^{V\times H\times W}$ contains the corresponding confidence values.

\paragraph{Future geometry supervision.}
To supervise future-geometry prediction, we pass the ground-truth future images through the same geometry encoder in a target branch, obtaining geometry-token targets $\bar{\mathbf{X}}_{t+k}^{\ell}\in\mathbb{R}^{V\times P\times D}$.
We use these tokens as supervision for the predicted future features, and the future feature prediction loss is defined as:
\begin{equation}
    \mathcal{L}_{\mathrm{feat}}
    =
    \frac{1}{FL}
    \sum_{k=1}^{F}
    \sum_{\ell=1}^{L}
    \left(
    1-\operatorname{cos}
    \left(
    \hat{\mathbf{X}}_{t+k}^{\ell},
    \operatorname{sg}\!\left(\bar{\mathbf{X}}_{t+k}^{\ell}\right)
    \right)
    \right),
    \label{eq:feature_distillation}
\end{equation}
where $\operatorname{sg}(\cdot)$ denotes stop-gradient and $\operatorname{cos}(\cdot,\cdot)$ averages cosine similarity over cameras and patch locations.
The predicted point maps are additionally supervised by dense future point-map targets $\mathbf{P}_{t+1:t+F}$ and their validity masks generated by MoGe-2~\citep{NEURIPS2025_336572db}.
The point-map objective, averaged over future steps, views, and valid pixels, combines Euclidean point regression, confidence-aware regression, and multi-scale surface-normal consistency:
\begin{equation}
    \mathcal{L}_{\mathrm{point}}^{\mathrm{future}}
    =
    \mathcal{L}_{\mathrm{reg}}
    +
    \mathcal{L}_{\mathrm{conf}}
    +
    \mathcal{L}_{\mathrm{normal}}.
    \label{eq:future_point_loss}
\end{equation}
We apply the same point-map objective to the encoded current frame, producing $\mathcal{L}_{\mathrm{point}}^{\mathrm{current}}$ and anchoring the geometry encoder while it learns to forecast.
The geometry pretraining objective is
\begin{equation}
    \mathcal{L}_{\mathrm{pre}}
    =
    \mathcal{L}_{\mathrm{feat}}
    +
    \mathcal{L}_{\mathrm{point}}^{\mathrm{future}}
    +
    \mathcal{L}_{\mathrm{point}}^{\mathrm{current}}.
    \label{eq:geometry_pretraining_loss}
\end{equation}

\subsection{GeoWAM: Visual Geometry World Action Model}
\label{sec:geowam_planning}

After geometry pretraining, the model is able to predict how the scene geometry will evolve from historical observations.
We extend this capability to trajectory planning through the action branch illustrated in \cref{fig:framework} (c).
Our design follows an inverse-dynamics-like pathway: the future pose decoder and trajectory head conditions on predicted future geometry to infer future trajectory.

\paragraph{Future ego-token decoding.}
We introduce $N_e$ learned ego-query seeds $\{\mathbf{q}_{n}^{\mathrm{ego}}\}_{n=1}^{N_e}$, with one seed for each ego-token slot.
For every future step $k$ and camera view $v$, the corresponding query is constructed as
\begin{equation}
    \mathbf{Q}_{t+k}^{\mathrm{ego},v,n}
    =
    \mathbf{q}_{n}^{\mathrm{ego}}
    +\mathbf{e}_{K+k}^{\mathrm{time}}
    +\mathbf{e}_{v}^{\mathrm{view}},
    \quad n=1,\ldots,N_e.
    \label{eq:future_ego_queries}
\end{equation}
The future pose decoder follows the same model structure as the future geometry decoder in \cref{fig:framework} (d).
Unlike the geometry decoder, its queries cross-attend to both the historical memory $\mathcal{Z}_{t}$ and the predicted future geometry tokens $\hat{\mathcal{U}}_{t+1:t+F}$, producing future ego tokens:
\begin{equation}
    \hat{\mathbf{E}}_{t+1:t+F}
    =
    \mathcal{D}_{\eta}^{\mathrm{ego}}
    \left(
    \mathbf{Q}^{\mathrm{ego}},
    \mathcal{Z}_{t},
    \operatorname{sg}\!\left(\hat{\mathcal{U}}_{t+1:t+F}\right)
    \right).
    \label{eq:future_ego_tokens}
\end{equation}
Here, $\hat{\mathbf{E}}_{t+k}\in\mathbb{R}^{V\times N_e\times D}$ denotes the predicted ego tokens at future step $t+k$.
The stop-gradient operation prevents the trajectory loss from propagating through the predicted future geometry, helping preserve the geometry forecasting capability acquired during pretraining.
This one-way connection also reflects our inverse-dynamics-like design.

\paragraph{Trajectory decoding.}
The action head takes the deepest-level historical ego tokens $\mathbf{E}_{t-K+1:t}^{L}$ together with the predicted future ego tokens $\hat{\mathbf{E}}_{t+1:t+F}$.
It appends the two sequences along the temporal dimension and refines them with a causal temporal transformer, allowing each future ego token to incorporate the preceding historical and predicted motion context.
A learned trajectory query then cross-attends to the refined historical and future ego features, and a regression head maps the resulting query to a single future trajectory.

\paragraph{Planning objective.}
During planning finetuning, we retain the future and current geometry objectives and supervise the trajectory with an $\ell_1$ regression loss $\mathcal{L}_{\mathrm{traj}}$.
We additionally predict the relative poses between historical frames using an auxiliary $\ell_1$ loss $\mathcal{L}_{\mathrm{pose}}$.
The complete finetuning objective is
\begin{equation}
    \mathcal{L}_{\mathrm{plan}}
    =
    \mathcal{L}_{\mathrm{pre}}
    +
    \lambda_{\mathrm{traj}}\mathcal{L}_{\mathrm{traj}}
    +
    \lambda_{\mathrm{pose}}\mathcal{L}_{\mathrm{pose}}.
    \label{eq:planning_loss}
\end{equation}

\section{Experiments}
\label{sec:results}
\begin{table*}[htbp]
\centering
\vspace{-1em}
\caption{\textbf{Future geometry prediction performance on nuScenes at different horizons.}
PAI denotes the NVIDIA PhysicalAI-Autonomous-Vehicles dataset~\citep{nvidia2025physicalai}.
Bold and underlined values indicate the best and second-best results, respectively.}
\label{tab:future_geometry}
\resizebox{\textwidth}{!}{
\begin{tabular}{l|ccccc|ccccc}
\toprule
& \multicolumn{5}{c|}{\textbf{Abs Rel} $\downarrow$}
& \multicolumn{5}{c}{\boldmath$\delta < 1.25$ $\uparrow$} \\
\cmidrule(lr){2-6} \cmidrule(lr){7-11}
\textbf{Method/Horizon}
& \textbf{1s} & \textbf{2s} & \textbf{3s} & \textbf{4s} & \textbf{mean}
& \textbf{1s} & \textbf{2s} & \textbf{3s} & \textbf{4s} & \textbf{mean} \\
\midrule
Epona~\citep{zhang2025epona} \scriptsize + DVGT~\citep{zuo2025dvgt}
& 0.229 & 0.263 & 0.292 & 0.310 & 0.274
& 0.732 & 0.677 & 0.620 & 0.589 & 0.655 \\

Cosmos~3~\citep{agarwal2026cosmos} \scriptsize + DVGT~\citep{zuo2025dvgt}
& 0.300 & 0.376 & 0.405 & 0.422 & 0.376
& 0.588 & 0.513 & 0.464 & 0.447 & 0.503 \\

VGGT-World~\citep{sun2026vggtworld}
& 0.272 & 0.329 & 0.342 & 0.357 & 0.325
& 0.612 & 0.553 & 0.513 & 0.497 & 0.544 \\

\midrule
\rowcolor{gray!15}
\textbf{GeoWAM (Ours) \scriptsize w/o PAI pretraining}
& \underline{0.171} & \underline{0.190} & \underline{0.227} & \textbf{0.245} & \underline{0.208}
& \underline{0.849} & \underline{0.813} & \underline{0.777} & \textbf{0.716} & \underline{0.789} \\
\rowcolor{gray!15}
\textbf{GeoWAM (Ours) \scriptsize w/ PAI pretraining}
& \textbf{0.146} & \textbf{0.176} & \textbf{0.208} & \underline{0.276} & \textbf{0.202}
& \textbf{0.873} & \textbf{0.835} & \textbf{0.797} & \underline{0.704} & \textbf{0.802} \\
\bottomrule
\end{tabular}
}
\end{table*}

\subsection{Experimental Setup}
\label{sec:experimental_setup}
\paragraph{Datasets.}
For geometry pretraining, we use image sequences from the NAVSIM \texttt{navtrain} split~\citep{dauner2024navsim} and the training set of the NVIDIA PhysicalAI-Autonomous-Vehicles dataset (PAI)~\citep{nvidia2025physicalai}.
We evaluate future-geometry prediction on the nuScenes validation set~\citep{caesar2020nuscenes}.
For planning, we finetune GeoWAM on the NAVSIM \texttt{navtrain} split and evaluate it on the open-loop \texttt{navtest} benchmark~\citep{dauner2024navsim} and the more challenging, two-stage pseudo-closed-loop \texttt{navhard} benchmark~\citep{cao2025pseudosimulation}.
We additionally evaluate zero-shot planning on nuScenes to showcase the generalizability of our method.

\paragraph{Metrics.}
For future geometry prediction, we convert each predicted future point map to ray depth, and report absolute relative error and threshold accuracy $\delta<1.25$.
For planning, we use the Extended Predictive Driver Model Score (EPDMS) from NAVSIM v2.
For nuScenes planning, we report L2 displacement error and collision rate at 1, 2, and 3 seconds, together with their averages.

\subsection{Future Geometry Prediction}
\label{sec:future_geometry_evaluation}
Table~\ref{tab:future_geometry} compares GeoWAM with the geometry world model VGGT-World~\citep{sun2026vggtworld} and the video world models Epona~\citep{zhang2025epona} and Cosmos~3~\citep{agarwal2026cosmos} on nuScenes.
To ensure a fair comparison, none of the evaluated models is trained on nuScenes. For Epona, we use the checkpoint trained on NAVSIM \texttt{navtrain}.
For the video baselines, we reconstruct geometry from their predicted RGB frames using DVGT~\citep{zuo2025dvgt}.
Both GeoWAM variants achieve better Abs Rel and $\delta<1.25$ than every baseline at every horizon.
Geometry pretraining on PhysicalAI is most effective at short horizons, lowering Abs Rel from 0.171 to 0.146 at 1\,s and from 0.227 to 0.208 at 3\,s, while the variant without pretraining remains slightly better at the longest 4\,s horizon.
These results show that GeoWAM outperforms both geometry world model and geometry reconstructed from video world model.

\subsection{Planning on NAVSIM v2}
\label{sec:planning_evaluation}
\noindent\textbf{\texttt{navtest}.}
\Cref{tab:nav_v2} compares GeoWAM with perception-based planners and recent world-action models on the \texttt{navtest} split.
Without geometry pretraining, GeoWAM achieves an EPDMS of \textbf{90.2}.
Geometry pretraining on PhysicalAI further improves the score to \textbf{90.7}.

\begin{table}[h]
\centering
\vspace{-1em}
\caption{\textbf{Open-loop planning results on the NAVSIM v2 \texttt{navtest} split.}}
\label{tab:nav_v2}
\resizebox{\textwidth}{!}{
\begin{tabular}{l|ccccccccc|c}
\toprule
\textbf{Method} & \textbf{NC} $\uparrow$ & \textbf{DAC} $\uparrow$ & \textbf{DDC} $\uparrow$ & \textbf{TLC} $\uparrow$ & \textbf{EP} $\uparrow$ & \textbf{TTC} $\uparrow$ & \textbf{LK} $\uparrow$ & \textbf{HC} $\uparrow$ & \textbf{EC} $\uparrow$ & \textbf{EPDMS} $\uparrow$ \\
\midrule

Transfuser~\citep{chitta2022transfuser} & 96.9 & 89.9 & 97.8 & 99.7 & 87.1 & 95.4 & 92.7 & \underline{98.3} & 87.2 & 84.0 \\

Hydra-MDP++~\citep{li2025hydra} & 97.2 & 97.5 & 99.4 & 99.6 & 83.1 & 96.5 & 94.4 & 98.2 & 70.9 & 81.4 \\

DriveSuprim~\citep{yao2026drivesuprim} & 97.5 & 96.5 & 99.4 & 99.6 & \textbf{88.4} & 96.6 & 95.5 & \underline{98.3} & 77.0 & 83.1 \\

ARTEMIS~\citep{feng2025artemis} & 98.3 & 95.1 & 98.6 & \underline{99.8} & 81.5 & 97.4 & 96.5 & \underline{98.3} & -- & 83.1 \\

DiffusionDrive~\citep{liao2025diffusiondrive} & 98.2 & 96.2 & 99.5 & \underline{99.8} & 87.4 & 97.3 & 96.9 & \textbf{98.4} & 87.7 & 88.2 \\
WoTE~\citep{li2025end} & 98.5 & 96.8 & 98.8 & \underline{99.8} & 86.1 & 97.9 & 95.5 & \underline{98.3} & 82.9 & 87.7 \\
DriveVLA-W0~\citep{li2026drivevla}
                     & 98.4 & 95.2 & 99.4 & \textbf{99.9} & 86.6 & 97.9 & 97.8 & \underline{98.3} & 82.7 & 86.9 \\
PWM~\citep{zhao2026forecasting}
                     & 98.8 & 95.9 & 99.4 & \textbf{99.9} & 86.4 & \underline{98.4} & 97.6 & \underline{98.3} & 85.3 & 88.2 \\
DriveLaW~\citep{xia2026drivelaw}
                     & 98.7 & 96.9 & \underline{99.6} & \underline{99.8} & 87.5 & 98.3 & 97.6 & \textbf{98.4} & 77.4 & 88.6 \\
DVGT-2~\citep{zuo2026dvgt}
                     & 98.7 & \textbf{97.9} & \textbf{99.7} & \textbf{99.9} & \underline{87.9} & 98.0 & \textbf{98.2} & 98.2 & 77.0 & 89.6 \\
EponaV2~\citep{xu2026eponav2}
                     & 98.5 & 97.4 & 99.5 & \textbf{99.9} & \underline{87.9} & 98.1 & 97.7 & 98.2 & 77.4 & 88.9 \\
\midrule
\rowcolor{gray!15}\textbf{GeoWAM (Ours) \scriptsize w/o PAI pretraining}    
                     & 98.7 & \underline{97.7} & \textbf{99.7} & \textbf{99.9} & 87.0 & 98.1 & 97.9 
                     & \underline{98.3} & 86.8 & \underline{90.2}
\\
\rowcolor{gray!15}\textbf{GeoWAM (Ours) {\scriptsize w/ PAI pretraining}}    
                     & \textbf{99.0} & \textbf{97.9} & \textbf{99.7} & \textbf{99.9} & 86.7 & \textbf{98.5} & 97.9 
                     & \underline{98.3} & \underline{87.8} & \textbf{90.7}
\\
\bottomrule
\end{tabular}
}
\end{table}

\begin{table*}[htbp]
    \centering
    \caption{\textbf{\texttt{navhard} leaderboard.} Methods trained with reinforcement learning or PDMS-score supervision are shown in gray and marked with $\dagger$. Among the remaining methods, bold and underlined values indicate the best and second-best results, respectively.}
    \label{tab:sota_transposed}
    \small
    \setlength{\tabcolsep}{5pt}
    \resizebox{\textwidth}{!}{%
        \begin{tabular}{lc|ccccccccc|c}
            \toprule
            \textbf{Method} & \textbf{Stage}
            & \textbf{NC} $\uparrow$
            & \textbf{DAC} $\uparrow$
            & \textbf{DDC} $\uparrow$
            & \textbf{TLC} $\uparrow$
            & \textbf{EP} $\uparrow$
            & \textbf{TTC} $\uparrow$
            & \textbf{LK} $\uparrow$
            & \textbf{HC} $\uparrow$
            & \textbf{EC} $\uparrow$
            & \textbf{EPDMS} $\uparrow$ \\
            \midrule
            \multirow{2}{*}{DriveVLA-W0~\citep{li2026drivevla}}
            & S1 & 96.8 & 83.3 & 99.0 & 99.6 & \underline{84.6} & 95.3 & \underline{96.4} & 97.6 & 78.2 & \multirow{2}{*}{24.4} \\
            & S2 & 76.8 & 64.3 & 79.9 & 98.3 & 89.2 & 75.0 & 46.8 & 95.8 & 53.1 & \\
            \cmidrule(l){2-12}
            \multirow{2}{*}{DriveLaW~\citep{xia2026drivelaw}}
            & S1 & 97.3 & 89.1 & \underline{99.2} & 99.6 & 84.3 & \underline{97.1} & 96.2 & \textbf{97.8} & 67.6 & \multirow{2}{*}{30.6} \\
            & S2 & 82.5 & 67.6 & 83.5 & 98.1 & 84.8 & 78.5 & 45.8 & 96.4 & 57.3 & \\
            \cmidrule(l){2-12}
            \multirow{2}{*}{DVGT-2~\citep{zuo2026dvgt}}
            & S1 & 97.2 & 91.3 & 98.4 & \underline{99.8} & \textbf{84.8} & 95.5 & 95.5 & 97.5 & 71.4 & \multirow{2}{*}{31.7} \\
            & S2 & 77.8 & 73.8 & 81.3 & 98.3 & \textbf{91.5} & 73.2 & 48.0 & 83.9 & 45.1 & \\
            \cmidrule(l){2-12}
            \multirow{2}{*}{LTFv6~\citep{nguyen2026lead}}
            & S1 & 96.5 & 86.6 & \underline{99.2} & 99.5 & 84.4 & 95.1 & 94.4 & \underline{97.7} & 76.4 & \multirow{2}{*}{31.9} \\
            & S2 & 79.8 & 75.5 & 86.2 & 97.8 & \underline{89.5} & 76.0 & \underline{50.0} & 95.2 & 66.7 & \\
            \midrule
            \multirow{2}{*}{\color{gray}NavFormer$^{\dagger}$~\citep{cao2025pseudosimulation}}
            & \color{gray}S1 & \color{gray}96.2 & \color{gray}92.4 & \color{gray}95.7 & \color{gray}99.6 & \color{gray}83.8 & \color{gray}96.0 & \color{gray}94.7 & \color{gray}96.4 & \color{gray}60.9 & \multirow{2}{*}{\color{gray}34.1} \\
            & \color{gray}S2 & \color{gray}85.7 & \color{gray}81.0 & \color{gray}83.5 & \color{gray}97.6 & \color{gray}90.1 & \color{gray}82.4 & \color{gray}48.2 & \color{gray}94.9 & \color{gray}48.4 & \\
            \cmidrule(l){2-12}
            \multirow{2}{*}{\color{gray}EponaV2$^{\dagger}$~\citep{xu2026eponav2}}
            & \color{gray}S1 & \color{gray}97.3 & \color{gray}90.7 & \color{gray}99.4 & \color{gray}100.0 & \color{gray}83.3 & \color{gray}97.3 & \color{gray}97.3 & \color{gray}97.6 & \color{gray}60.9 & \multirow{2}{*}{\color{gray}36.1} \\
            & \color{gray}S2 & \color{gray}83.6 & \color{gray}78.0 & \color{gray}88.0 & \color{gray}98.9 & \color{gray}86.0 & \color{gray}80.3 & \color{gray}50.1 & \color{gray}96.1 & \color{gray}52.0 & \\
            \cmidrule(l){2-12}
            \multirow{2}{*}{\color{gray}RAP-DINO$^{\dagger}$~\citep{feng2026rap}}
            & \color{gray}S1 & \color{gray}97.1 & \color{gray}94.4 & \color{gray}98.8 & \color{gray}99.8 & \color{gray}83.9 & \color{gray}96.9 & \color{gray}94.7 & \color{gray}96.4 & \color{gray}66.2 & \multirow{2}{*}{\color{gray}36.9} \\
            & \color{gray}S2 & \color{gray}83.2 & \color{gray}83.9 & \color{gray}87.4 & \color{gray}98.0 & \color{gray}86.9 & \color{gray}80.4 & \color{gray}52.3 & \color{gray}95.2 & \color{gray}52.4 & \\
            \midrule
            \rowcolor{gray!15}
            & S1 & \underline{97.7} & \underline{91.5} & 99.1 & \underline{99.8} & 83.8 & 95.8 & 96.0 & \textbf{97.8} & 79.0 & \\
            \rowcolor{gray!15}\multirow{-2}{*}{\textbf{GeoWAM (Ours) \scriptsize w/o PAI pretraining}}
            & S2 & 80.4 & \textbf{76.3} & \underline{87.3} & \textbf{98.7} & 88.9 & 76.2 & 49.9 & 94.0 & 56.0 & \multirow{-2}{*}{\underline{36.6}} \\
            \cmidrule(l){2-12}
            \rowcolor{gray!15}
            & S1 & \textbf{98.9} & \textbf{92.9} & \textbf{99.6} & \textbf{100.0} & 82.0 & \textbf{97.3} & \textbf{96.9} & \textbf{97.8} & \textbf{79.5} & \\
            \rowcolor{gray!15}\multirow{-2}{*}{\textbf{GeoWAM (Ours) \scriptsize w/ PAI pretraining}}
            & S2 & \textbf{84.1} & \underline{76.0} & \textbf{87.4} & \underline{98.6} & 84.0 & \underline{80.2} & \textbf{50.5} & 96.9 & 63.7 & \multirow{-2}{*}{\textbf{39.6}} \\
            \bottomrule
        \end{tabular}%
    }
\end{table*}

\noindent\textbf{Two-Stage Planning on \texttt{navhard}.}
We further evaluate GeoWAM under the two-stage pseudo-closed-loop planning protocol on the \texttt{navhard} split~\citep{cao2025pseudosimulation}.
Importantly, GeoWAM \textbf{\textit{uses no PDMS-derived supervision}} during training, including reinforcement learning (RL) or a learned PDMS scorer.
As reported in \cref{tab:sota_transposed}, GeoWAM without PAI pretraining achieves an EPDMS of \textbf{36.6}, outperforming all baselines trained without PDMS supervision, as well as two baselines trained with PDMS supervision~\citep{cao2025pseudosimulation, xu2026eponav2}.
It trails only RAP~\citep{feng2026rap} (36.9), which is trained with PDMS-score supervision.
Geometry pretraining on PAI further raises the score by 3.0 points to \textbf{39.6}, surpassing all baselines, including RAP.

Together, these results show that future-geometry modeling provides an effective representation for downstream planning.
The larger pretraining gain on \texttt{navhard} further suggests that learned geometric dynamics are especially valuable when planning errors influence subsequent observations.

\subsection{Zero-shot Planning on NuScenes}
\label{sec:zero_shot_nuscenes}
To assess cross-dataset generalization and the effect of geometry pretraining, we evaluate two GeoWAM variants on nuScenes without nuScenes-specific trajectory supervision.
Both planners are trained only on NAVSIM \texttt{navtrain}. The pretrained variant additionally uses image sequences from \texttt{navtrain} and PhysicalAI for geometry pretraining.
As shown in Table~\ref{tab:plan_nusc_main}, even without geometry pretraining on PAI, GeoWAM achieves an average L2 error of 1.28\,m and collision rate of 0.24\%, outperforming both zero-shot video-based world models (\textit{i.e. }PWM and DriveVLA-W0).

\begin{table*}[h]
\caption{\textbf{Zero-shot motion planning performance on nuScenes.}}
\centering
\resizebox{\linewidth}{!}{
\begin{tabular}{l|c|c|cccc|cccc}
\toprule
\multirow{2}{*}{\textbf{Method}} & \multirow{2}{*}{\shortstack{\textbf{nuScenes}\\\textbf{Finetune}}} & \multirow{2}{*}{\textbf{Auxiliary Supervision}} &
\multicolumn{4}{c|}{\textbf{L2 (m)} $\downarrow$} & 
\multicolumn{4}{c}{\textbf{Collision Rate (\%)} $\downarrow$} \\
&&& \textbf{1s} & \textbf{2s} & \textbf{3s} & \cellcolor{gray!30}\textbf{Avg.} & \textbf{1s} & \textbf{2s} & \textbf{3s} & \cellcolor{gray!30}\textbf{Avg.}  \\
\midrule
ST-P3~\citep{hu2022st} & \cmark & Map\&Box\&Depth & 1.33 & 2.11 & 2.90 & \cellcolor{gray!30}2.11 & 0.23 & 0.62 & 1.27 & \cellcolor{gray!30}0.71  \\
UniAD~\citep{hu2023planning} & \cmark & {\footnotesize Map\&Box\&Motion} & {0.48} & {0.96} & {1.65} & \cellcolor{gray!30}{1.03} & {0.05} & 0.17 & 0.71 & \cellcolor{gray!30}0.31  \\
OccNet~\citep{tong2023scene} & \cmark & 3D-Occ\&Map\&Box & 1.29 & 2.13 & 2.99 & \cellcolor{gray!30}2.14 & 0.21 & 0.59 & 1.37 & \cellcolor{gray!30}0.72  \\
OccWorld~\citep{zheng2024occworld} & \cmark & 3D-Occ & 0.52 & 1.27 & 2.41 & \cellcolor{gray!30}1.40 & 0.12 & 0.40 & 2.08 & \cellcolor{gray!30}0.87  \\
VAD-Tiny~\citep{jiang2023vad} & \cmark & Map\&Box\&Motion & 0.60 & 1.23 & 2.06 & \cellcolor{gray!30}1.30 & 0.31 & 0.53 & 1.33 & \cellcolor{gray!30}0.72  \\
VAD-Base~\citep{jiang2023vad} & \cmark & Map\&Box\&Motion & 0.54 & 1.15 & 1.98 & \cellcolor{gray!30}1.22 & 0.04 & 0.39 & 1.17 & \cellcolor{gray!30}0.53 \\
GenAD~\citep{zheng2024genad} & \cmark & Map\&Box\&Motion & \underline{0.36} & \underline{0.83} & \textbf{1.55} & \cellcolor{gray!30}\underline{0.91} & 0.06 & {0.23} & {1.00} & \cellcolor{gray!30}{0.43} \\
Doe-1~\citep{zheng2024doe} & \cmark & QA & 0.50 & 1.18 & 2.11 & \cellcolor{gray!30}1.26 & 0.04 & 0.37 & 1.19 & \cellcolor{gray!30}0.53  \\
Epona~\citep{zhang2025epona} & \cmark & Future RGB & 0.61 & 1.17 & 1.98 & \cellcolor{gray!30}1.25 & \underline{0.01} & 0.22 & 0.85 & \cellcolor{gray!30}0.36  \\
\midrule
PWM~\citep{zhao2026forecasting} & \textbf{\xmark} & Future RGB & 2.06 & 3.91 & 6.00 & \cellcolor{gray!30}3.99 & 0.12 & 0.15 & 0.86 & \cellcolor{gray!30}0.36  \\
DriveVLA-W0~\citep{li2026drivevla} & \textbf{\xmark} & Future RGB & 0.43 & 1.26 & 2.60 & \cellcolor{gray!30}1.43 & 0.22 & 0.66 & 1.42 & \cellcolor{gray!30}0.77  \\
\textbf{GeoWAM (Ours) \scriptsize w/o PAI pretraining} & \textbf{\xmark} & Future Geometry & 0.42 & 1.19 & 2.24 & \cellcolor{gray!30}1.28 & \textbf{0.00} & \textbf{0.10} & \underline{0.60} & \cellcolor{gray!30}\underline{0.24}  \\
\textbf{GeoWAM (Ours) \scriptsize w/ PAI pretraining} & \textbf{\xmark} & Future Geometry & \textbf{0.29} & \textbf{0.79} & \underline{1.58} & \cellcolor{gray!30}\textbf{0.89} & 0.02 & \underline{0.12} & \textbf{0.23} & \cellcolor{gray!30}\textbf{0.12}  \\
\bottomrule
\end{tabular}%
}
\label{tab:plan_nusc_main}
\end{table*}

Notably, this variant already attains a lower collision rate than every baseline in the table, including planners finetuned on nuScenes, even though its L2 error still trails the strongest finetuned planners.
Geometry pretraining further reduces the averages to 0.89\,m and 0.12\%, respectively, without additional trajectory supervision, surpassing all the nuScenes-finetuned baselines on both average metrics, suggesting that future geometry supports cross-dataset planning and that geometry pretraining strengthens this transfer.

\subsection{Scaling Geometry Pretraining}
\label{sec:geometry_pretraining_scaling}
Our first-stage geometry pretraining (\cref{sec:geometry_world_model}) learns future scene geometry directly from image sequences, allowing it to leverage additional driving data without ground-truth trajectory annotations.
To evaluate this scalability, we vary the amount of PhysicalAI data used for geometry pretraining from 0 to 832 hours, with 832 hours covering the entire PhysicalAI training set.
The left panel of \cref{tab:geometry_pretraining_scaling} shows that scaling the pretraining data improves \texttt{navhard} EPDMS from 36.6 to 39.6 and reduces the nuScenes zero-shot collision rate from 0.24\% to 0.12\%.
The right panel shows the corresponding relative improvements over the model without geometry pretraining.
The nuScenes L2 error is not strictly monotonic, since it measures deviation from the recorded human trajectory and may penalize other valid driving behaviors.
In contrast, the collision rate decreases consistently as the pretraining data scales, indicating improved planning safety.
As shown by the parameter comparison in appendix, GeoWAM uses 1.9B parameters, substantially fewer than DriveVLA-W0 and EponaV2.
Its planning performance nevertheless continues to improve with additional geometry-pretraining data, without increasing model capacity or trajectory supervision.
\begin{table}[h]
    \centering
    \caption{\textbf{Effect of geometry-pretraining data scale on downstream planning.} Left: absolute planning performance at different pretraining-data scales. Right: relative improvements over the model without geometry pretraining on \texttt{navhard} and nuScenes zero-shot planning.}
    \label{tab:geometry_pretraining_scaling}
    \begin{minipage}[c]{0.53\linewidth}
    \centering
    \setlength{\tabcolsep}{4pt}
    \resizebox{\linewidth}{!}{%
    \begin{tabular}{c|ccc|cccc|cccc}
        \toprule
        \multirow{3}{*}{\shortstack{\textbf{PAI} \\ \textbf{Data} \\ \textbf{(hours)}}}
        & \multicolumn{3}{c|}{{\textbf{\texttt{navhard}}}}
        & \multicolumn{8}{c}{\textbf{nuScenes Zero-shot}} \\
        & \multicolumn{3}{c|}{\textbf{EPDMS} $\uparrow$}
        & \multicolumn{4}{c|}{\textbf{L2 (m)} $\downarrow$}
        & \multicolumn{4}{c}{\textbf{Collision Rate (\%)} $\downarrow$} \\
        & \textbf{S1}
        & \textbf{S2}
        & \cellcolor{gray!30}\textbf{Comb.}
        & \textbf{1s} & \textbf{2s} & \textbf{3s} & \cellcolor{gray!30}\textbf{Avg.}
        & \textbf{1s} & \textbf{2s} & \textbf{3s} & \cellcolor{gray!30}\textbf{Avg.} \\
        \midrule
        0 & 80.7 & 43.5 & \cellcolor{gray!30}36.6 & 0.42 & 1.19 & 2.24 & \cellcolor{gray!30}1.28 & \textbf{0.00} & \textbf{0.10} & 0.60 & \cellcolor{gray!30}0.24 \\
        100 & 81.3 & 45.5 & \cellcolor{gray!30}37.8 & 0.31 & 0.81 & 1.64 & \cellcolor{gray!30}0.92 & 0.02 & 0.12 & 0.44 & \cellcolor{gray!30}0.19  \\
        496 & 81.2 & 45.8 & \cellcolor{gray!30}38.2 & \textbf{0.29} & \textbf{0.76} & \textbf{1.55} & \cellcolor{gray!30}\textbf{0.86} & 0.04 & 0.12 & 0.33 & \cellcolor{gray!30}0.16 \\
        832 & \textbf{83.1} & \textbf{46.7} & \cellcolor{gray!30}\textbf{39.6} & \textbf{0.29} & 0.79 & 1.58 & \cellcolor{gray!30}0.89 & 0.02 & 0.12 & \textbf{0.23} & \cellcolor{gray!30}\textbf{0.12}  \\
        \bottomrule
    \end{tabular}
    }
    \par\vspace{2pt}
    \end{minipage}%
    \hfill
    \begin{minipage}[c]{0.46\linewidth}
        \centering
        \includegraphics[width=\linewidth]{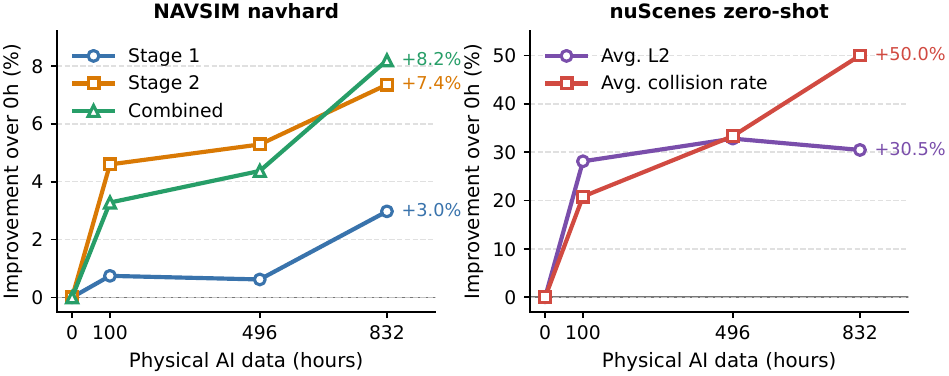}
        \par\vspace{2pt}
    \end{minipage}%
\end{table}

\vspace{-1em}
\subsection{Ablation Study}
\label{sec:ablation_study}
We ablate three factors in two groups (\cref{tab:planning_ablations}): future-geometry modeling and PAI pretraining in (a), and the prediction horizon in (b).
(b) is conducted without PAI pretraining.

\noindent\textbf{Future geometry and PAI pretraining.}
Removing future-geometry prediction while retaining current-frame geometry supervision lowers EPDMS from 90.2 to 89.2 on \texttt{navtest} and from 36.6 to 31.7 on \texttt{navhard}.
These gaps isolate the benefit of forecasting future geometric evolution.
PAI pretraining further improves EPDMS by 0.5 points on \texttt{navtest} and 3.0 points on \texttt{navhard}.
The larger gain on \texttt{navhard} suggests that pretraining on more diverse road layouts and traffic scenes strengthens the learned geometric dynamics, particularly when planning errors affect subsequent observations under pseudo-closed-loop evaluation.

\begin{table}[t]
  \centering
  \vspace{-1em}
  \caption{\textbf{Ablation studies on downstream planning.}}
  \label{tab:planning_ablations}
  \begin{minipage}[b]{0.72\linewidth}
  \centering
  \resizebox{\linewidth}{!}{%
  \begin{tabular}{l|ccc|cc}
  \toprule
  \textbf{Variant}
  & \shortstack{\textbf{Current}\\\textbf{Geometry}}
  & \shortstack{\textbf{Future}\\\textbf{Geometry}}
  & \shortstack{\textbf{PAI}\\\textbf{Pretraining}}
  & \shortstack{\textbf{\texttt{navtest}}\\\textbf{EPDMS} $\uparrow$}
  & \shortstack{\textbf{\texttt{navhard}}\\\textbf{EPDMS} $\uparrow$} \\
  \midrule
  w/o Future Geometry & \cmark & & & 89.2 & 31.7 \\
  w/o PAI Pretraining & \cmark & \cmark & & 90.2 & 36.6 \\
  \textbf{Full} & \cmark & \cmark & \cmark
  & \textbf{90.7} & \textbf{39.6} \\
  \bottomrule
  \end{tabular}%
  }
  \par\vspace{2pt}{\small (a) Future geometry and PAI pretraining}
  \end{minipage}%
  \hfill
  \begin{minipage}[b]{0.27\linewidth}
  \centering
  \resizebox{\linewidth}{!}{%
  \begin{tabular}{c|cc}
  \toprule
  \textbf{\# Frames}
  & \shortstack{\textbf{\texttt{navtest}}\\\textbf{EPDMS} $\uparrow$}
  & \shortstack{\textbf{\texttt{navhard}}\\\textbf{EPDMS} $\uparrow$} \\
  \midrule
  2 & 89.2 & 30.7 \\
  4 & 89.5 & 32.2 \\
  6 & 89.9 & 34.3 \\
  \textbf{8} & \textbf{90.2} & \textbf{36.6} \\
  \bottomrule
  \end{tabular}%
  }
  \par\vspace{2pt}{\small (b) Prediction horizon}
  \end{minipage}
  \end{table}

\noindent\textbf{Prediction horizon.}
With $F$ future frames predicted at 2\,Hz, the forecast horizon ranges from 1\,s ($F{=}2$) to the full 4\,s planning horizon ($F{=}8$).
EPDMS increases monotonically with $F$, by 1.0 on \texttt{navtest} and 5.9 on \texttt{navhard} in total, and the per-step gain on \texttt{navhard} grows rather than saturates (+1.5, +2.1, and +2.3).
In contrast, forecasting only 1\,s ahead brings no benefit: $F{=}2$ matches the variant without future geometry on \texttt{navtest} (89.2) and falls below it on \texttt{navhard} (30.7 vs.\ 31.7).
This suggests that the improvement does not stem from the additional decoder capacity alone, but from exposing scene dynamics over a horizon aligned with the planned trajectory.

\subsection{Ego-Motion Recovery from Predicted States}
\label{sec:ego_motion_recovery}
To examine how directly different predicted states expose future ego motion, we compare trajectories recovered from PWM's future video~\citep{zhao2026forecasting} and GeoWAM's future geometry.
For PWM, we recover ego motion from generated RGB frames using two separate methods: classical epipolar geometry~\citep{Hartley_Zisserman_2004} and the learning-based VGGT~\citep{wang2025vggt}.
For GeoWAM, we match static 3D keypoints across predicted point maps and recover the relative rigid transforms with Kabsch alignment~\citep{Kabsch:a12999}.

\begin{figure}[htbp]
    \centering
    \includegraphics[width=1\linewidth]{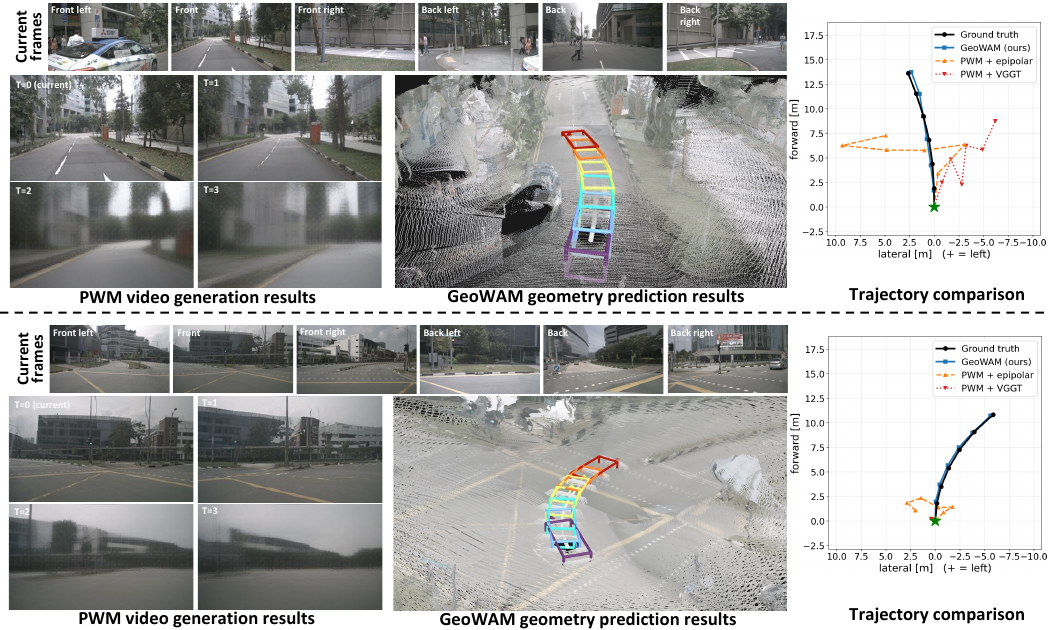}
    \caption{\textbf{Ego-motion recovery from predicted future states.} Each row shows current observations, PWM-generated future video, GeoWAM-predicted future geometry, and trajectories recovered from different future representation, compared with the logged ground truth.}
    \label{fig:visualization}
\end{figure}

We show two examples in \cref{fig:visualization}, including each model's predicted future state alongside the trajectories recovered by all three methods and the ground truth trajectory.
In these examples, the geometry-derived trajectories follow the ground truth more closely, whereas the trajectories recovered from future video show larger deviations.
This qualitative comparison illustrates how future geometry makes ego motion more directly accessible than pixel-space predictions.
More visualizations are provided appendix.

\section{Conclusion}
\label{sec:conclusion}

We introduced GeoWAM, a visual geometry world action model that treats geometry as the state space for autonomous driving.
Rather than forecasting future pixels, GeoWAM predicts future scene geometry from historical multiview observations and uses the learned geometric dynamics to condition trajectory prediction through an inverse-dynamics-like action branch.
Experiments on future-geometry prediction and planning under various settings consistently demonstrate the advantage of this geometry-based formulation over image-based alternatives.
Scaling geometry pretraining with additional driving data further improves planning performance, without increasing model capacity or trajectory supervision.
These results establish geometry as an effective, planning-aligned state representation and geometry pretraining as a general, scalable strategy for downstream planning.

\bibliographystyle{plainnat}
\bibliography{main}

\end{document}